\documentclass[runningheads]{llncs}
\usepackage[T1]{fontenc}
\usepackage{graphicx,verbatim}
\usepackage{amsmath,amssymb}
\usepackage{amsfonts}
\usepackage{mathtools}

\usepackage{lipsum}
\usepackage{float}
\usepackage{mathrsfs, bm}
\DeclareMathAlphabet{\mathcal}{OMS}{cmsy}{m}{n}
\DeclareSymbolFont{largesymbols}{OMX}{cmex}{m}{n}
\usepackage{afterpage}
\usepackage{caption}
\usepackage{subfigure}
\usepackage{color,xcolor}
\definecolor{gre}{RGB}{50, 168, 82}
\usepackage{tabularx}
\usepackage{multirow}
\usepackage{pifont}
\usepackage{array}
\usepackage{hyperref}
\usepackage{threeparttable}
\usepackage{bbding}
\hypersetup{
    colorlinks=true,
    linkcolor=black,
    citecolor=blue,
    urlcolor=blue,
    linkbordercolor=white,
}
\newcolumntype{C}[1]{>{\centering\arraybackslash}p{#1}}
\begin{document}
\title{Blind Restoration of High-Resolution Ultrasound Video}
%
\author{Chu Chen\inst{1,2}\textsuperscript{\Envelope} \and
Kangning Cui\inst{1,2} \and
Pasquale Cascarano\inst{3} \and
Wei Tang\inst{1,2} \and
Elena Loli Piccolomini\inst{4} \and
Raymond H. Chan\inst{2,5}}

\authorrunning{C. Chen et al.}
%
\institute{Department of Mathematics, City University of Hong Kong, Hong Kong \and
Hong Kong Centre for Cerebro-cardiovascular Health Engineering, Hong Kong\\ \and
Department of the Arts, University of Bologna, Italy\\ \and
Department of Computer Science and Engineering, University of Bologna, Italy\\ \and
Department of Operations and Risk Management and School of Data Science, Lingnan University, Hong Kong \\
\email{chuchen4-c@my.cityu.edu.hk}
}

\maketitle              
\begin{abstract}
Ultrasound imaging is widely applied in clinical practice, yet ultrasound videos often suffer from low signal-to-noise ratios (SNR) and limited resolutions, posing challenges for diagnosis and analysis. Variations in equipment and acquisition settings can further exacerbate differences in data distribution and noise levels, reducing the generalizability of pre-trained models. This work presents a self-supervised ultrasound video super-resolution algorithm called Deep Ultrasound Prior (DUP). DUP employs a video-adaptive optimization process of a neural network that enhances the resolution of given ultrasound videos without requiring paired training data while simultaneously removing noise. Quantitative and visual evaluations demonstrate that DUP outperforms existing super-resolution algorithms, leading to substantial improvements for downstream applications.

\keywords{Ultrasound \and Video Super-resolution \and Deep Image Prior \and Self-supervised Learning \and Ejection Fraction.}

\end{abstract}

\section{Introduction}

Ultrasound (US) imaging is an essential diagnostic tool in modern medicine, widely utilized in various clinical applications, including cardiology~\cite{aly2021cardiac,pellicori2021ultrasound}, obstetrics~\cite{leung2021applications,alfirevic2015fetal}, and musculoskeletal~\cite{patil2012role,van1999ultrasound} imaging. Its ability to provide visualization of internal structures, coupled with its non-invasive nature, makes US a preferred imaging technique. Furthermore, the dynamic recording of US videos captures much more information about tissues than single images, making diagnosis heavily reliant on video analysis. Applications include breast lesion detection~\cite{yoon2016validation,zhao2023deep} and cardiovascular monitoring~\cite{duffy2022high,christensen2024vision,tang2024bilateral}, where metrics like Ejection Fraction (EF) play a critical role. However, the quality of US data often suffers from both inherent and technical limitations. Acoustic noise, caused by scattering and absorption of sound waves in complex tissue structures, degrades image clarity, while limited resolution, stemming from constraints in transducer technology and frequency ranges, affects the ability to capture fine anatomical details.
These challenges not only affect diagnostic accuracy but also impact downstream tasks like EF prediction, where precise visualization of cardiac structures and motion is essential. Addressing these limitations requires continuous advancements in US technology to enhance visual quality and diagnostic reliability through super-resolution (SR) algorithms.

\subsubsection{Related Works.}

The development of large training datasets has significantly advanced the research on SR algorithms for natural images~\cite{agustsson2017ntire} and videos~\cite{nah2019ntire,liu2013bayesian}.
These datasets enable supervised training of neural networks, resulting in models that achieve exceptional performance. Among all, deep convolutional neural networks (CNN)~\cite{lim2017enhanced,zhang2018residual} have shown remarkable capabilities in single image SR (SISR) tasks. To ensure temporal correlation in video enhancement, techniques such as the flow-based~\cite{Revaud_2015_CVPR,xue2019video} and attention-based method~\cite{zhou2024video} are integrated with CNNs. However, significant shifts in data distributions make these pre-trained models struggle with generalization. Consequently, artifacts are often introduced in the upsampled videos.

In the field of US image SR, acquiring paired low-resolution (LR) and high-resolution (HR) datasets is particularly challenging, leading researchers to either create simulated datasets for supervised training or explore unsupervised/self-supervised approaches. Choi et al.~\cite{Choi2018IUS} modified the SRGAN~\cite{Ledig_2017_CVPR} to enhance B-mode US images with low lateral resolution, improving their similarity to HR images. A U-net-style network~\cite{van2019deep} is trained on simulated pairs for super-resolved US images. Liu et al.~\cite{liu2021perception} proposed ZSSR-Cycle, a zero-shot and self-supervised generative adversarial network framework for US image SR.

One particularly intriguing development in image processing is the Deep Image Prior (DIP) method~\cite{Ulyanov_2018_CVPR}, which demonstrates that CNN structures can inherently capture and enhance image features without relying on any predefined training dataset. In fact, it has been shown that CNNs can more effectively replicate and enhance the structural characteristics present in images compared to arbitrary noise patterns. In the field of SISR, DIP has achieved competitive performance compared to other supervised training models. Variations of DIP have also been proposed for video processing. The Deep Video Prior (DVP)~\cite{lei2022deep} adopted DIP for individual video frame with iteratively reweighted training strategy to address the multi-modal inconsistencies. The Recursive Deep Prior Video (RDPV)~\cite{cascarano2021recursive} proposed a recursive updating rule for CNN optimization when dealing with each frame for the light Time-Lapse Microscopy video SR. While DIP-based methods effectively capture underlying image structures, their ability to fit fine details can also lead to overfitting, where noise and artifacts are misinterpreted as part of the high-frequency details that SR methods seek to enhance. These unwanted patterns adversely affect diagnostic and analytical applications based on the SR-enhanced US videos.

\subsubsection{Contributions.} US video restoration faces fundamental challenges from lacking paired data, heterogeneous distributions, and multi-factorial corruptions. We address these through a blind restoration method followed by extensive evaluations. The key contributions of this work are as follows:
\begin{itemize}
    \item We present a self-supervised framework Deep Ultrasound Prior (DUP) for US video SR, eliminating the need for paired training data through video-specific optimization.
    \item DUP adopts a Weight Inheritance (WIn) strategy and dual regularization for CNN optimization, accelerating convergence and achieving information sharing among successive frames while removing noise and artifacts that appeared in US videos.
    \item We systematically compare DUP with existing SR techniques under various levels of noise, which demonstrates the superior performance and robustness of our approach in US video restoration.
    \item By evaluating EF prediction from restored cardiac videos, DUP achieves the lowest error, highlighting its potential to improve clinical diagnostic accuracy.
\end{itemize}

\section{Method}
\begin{figure}[!t]
\centering
  \includegraphics[width=\linewidth]{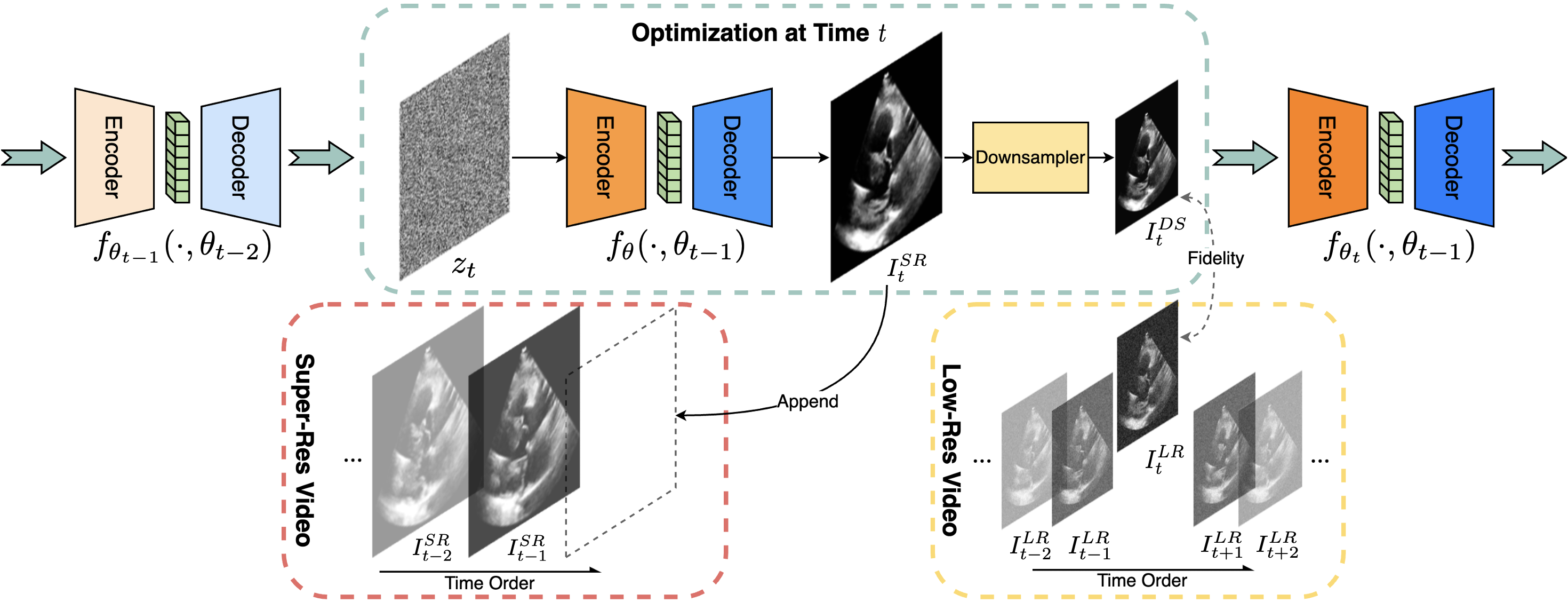}
  \caption{Schematic of the DUP algorithm.}
  \label{flowchart}
\end{figure}
\subsection{Problem Formulation}
The inverse problem of frame-wise video SR can be formulated as 
\begin{equation}
    I_t^{LR} = \mathcal{A}I_t^{HR} + e_t, \quad t=1,2,\dots,T,
    \label{prob}
\end{equation}
where $\mathcal{A}$ represents the measurement operator, $I_t^{LR} \in \mathbb{R}^{MN}$ is the vectorized $t$th LR video frame with spatial resolution of $M\times N$, $T$ is the number of frames, $I_t^{HR} \in \mathbb{R}^{s^2MN}$ is the underlying HR frame with upscaling factor $s \in \mathbb{Z}^+$, and $e_t \in \mathbb{R}^{MN}$ is the measurement error. In this work, we focus on the linear degradation operator $\mathcal{A} \in \mathbb{R}^{MN \times s^2MN}$. Since the SR problem is known to be strongly ill-posed~\cite{yue2016image}, the reconstructed SR frame $I_t^{SR}$ that solves this system is not unique while sensitive to the measurement error $e_t$. The common approach to address this problem is to add prior constraints on the desired $I_t^{SR}$, narrowing the space of the possible solution. We formulate the underdetermined systems in Eq.~(\ref{prob}) into a set of variational regularized optimization problems:
\begin{equation}
    I_t^{SR} = \operatorname*{argmin}_{I_t} \frac{1}{2}||I_t^{LR}-\mathcal{A}I_t||_2^2 + \lambda R(I_t),\quad t=1,2,\dots,T,
    \label{optim_prob}
\end{equation}
where $\{I_t\}_{t=1}^T$ are variables.
In Eq.~(\ref{optim_prob}), the SR video $\{I_t^{SR}\}_{t=1}^T$ is estimated by minimizing the sum of the $\ell_2$ fidelity term and the regularization term $R$ for each frame, where $\lambda$ is the regularization parameter that balances the two terms.

\subsection{Deep Ultrasound Prior}
The ability to reconstruct and augment detail highlights CNNs' potential for self-supervised learning techniques, improving the spatial resolution and clarity of US video. We now explain our proposed DUP method in detail.

This algorithm produces super-resolved US video in a frame-by-frame manner following the time order. Let integer $t\in[1, T]$ be an index of the video frames and $z_t\in\mathbb{R}^{s^2MN}$ be a random image with targeted spatial resolution. DUP utilizes an encoder-decoder CNN treated as a mapping from the random image $z_t$ to SR frame $I_t^{SR} = f_{\theta}(z_t,\omega): \mathbb{R}^{s^2MN}\times\mathbb{R}^\Theta \rightarrow \mathbb{R}^{s^2MN}$, where the sub-index $\theta$ is the trainable parameters of the CNN and $\Theta$ is the total number of parameters, the second entry $\omega \in \mathbb{R}^\Theta$ defined initialization of CNN.

During the reconstruction process of SR images, we aim to enhance only the contrast details while preventing the amplification of random noise and artifacts and even removing them. A widely recognized approach for achieving this is the total variation (TV) model~\cite{rudin1992nonlinear}, which replaces the regularization term $R$ in Eq.~(\ref{optim_prob}) with a TV regularization term $R_1$:
\begin{equation}
    R_1(I_t) = \sum_{i=1}^{s^2MN}(|(D_hI_t)_i| + |(D_vI_t)_i|)
\end{equation}
where $D_h$ and $D_v$ are the first-order finite difference discrete operators along the horizontal and vertical directions, respectively.

However, optimizing with only total variation regularization can result in staircase effects appearing in the image. To address the piece-wise constant artifacts, we additionally introduce a higher-order (HO)~\cite{chan2000high,bredies2010total,hintermuller2006infeasible} term $R_2$:
\begin{equation}
    R_2(I_t) = \frac{1}{2}\sum_{i=1}^{s^2MN}(|(D_hI_t)_i|^2 + |(D_vI_t)_i|^2).
\end{equation}

Hence, the optimization problem in Eq.~(\ref{optim_prob}) is solved by optimizing the parameters of the CNN as
\begin{equation}
    \theta_t = \operatorname*{argmin}_{\theta} \frac{1}{2}||I_t^{LR}-\mathcal{A}f_{\theta}(z_t,\omega)||_2^2 + \lambda_1 R_1(f_{\theta}(z_t,\omega)) + \lambda_2 R_2(f_{\theta}(z_t,\omega)),\\
    \label{optim_params}
\end{equation}
thereby the SR frames can be reconstructed by $I_t^{SR} = f_{\theta_t}(z_t,\omega)$.

The overview of the proposed DUP framework is shown in Fig.~\ref{flowchart}. We now explain in detail the technical advancement of DUP.

\subsubsection{WIn Strategy}For any frame-by-frame video enhancement algorithm, leveraging temporal correlation is essential for maintaining high-quality results. DUP employs a Weight Inheritance (WIn) strategy to utilize the continuity of adjacent frames for improved efficiency, allowing the CNN to converge faster while accumulating optimization iterations. The WIn strategy in DUP operates as follows: 1. The CNN is randomly initialized with parameter $\theta_0$ as processing the first video frame. After optimization based on the first low resolution frame $I_1^{LR}$, the optimal parameter of the network is $\theta_1$ and corresponding super-resolved image $I_1^{SR} = f_{\theta_1}(z_1,\theta_0)$; 2. For subsequent frames $I_t^{LR}$ ($t\geq 2$), CNN is initialized with the optimized parameters $\theta_{t-1}$ from the preceding frame $I_{t-1}^{LR}$, and the super-resolved image is generalized by $I_t^{SR} = f_{\theta_t}(z_t,\theta_{t-1})$. This WIn strategy enhances the computational efficiency in frame-by-frame video processing. By sharing network parameters, CNNs utilize information from neighboring frames, exploiting the temporal correlations and achieving faster convergence within a limited number of iterations. Furthermore, continuous weight inheritance means the CNN effectively trains on more samples as it processes additional frames, leading to better SR performance through extended iterations.

\subsubsection{Early Stopping and Input Update} Since DUP follows a self-supervised learning strategy, there is no need to consider generalization beyond the given video. However, excessive optimization can introduce unwanted artifacts into the image. Specifically, over-optimizing the fidelity term may lead to overfitting measurement errors, whereas excessive optimization of the regularization term can cause over-smoothing. To mitigate these issues, the iterative process is early-stopped based on a criterion that monitors the loss over a fixed window. Training stops when the loss value stabilizes. With the WIn strategy, the CNN typically requires fewer iterations after processing the first frame, allowing DUP to begin monitoring the loss after fewer training steps. 

Furthermore, according to DIP, the input to the CNN is a randomly generated noise image $z_1$. Due to the strong similarities and periodic patterns exhibited among frames in a US video, using the same noise image as input for every frame (i.e., $z_t=z_1,t\geq2$) may cause the CNN to undergo lazy training, leading to outputs that remain overly similar to the first frame. In response, DUP updates the noise image for each frame as follows: 1. $z_1 = n_1$; 2. $z_t = z_{t-1} + \sigma n_t, (t\geq 2)$, where each pixel of the random image $\{n_t\}_{t=1}^{T}\in\mathbb{R}^{s^2MN}$ is $i.i.d$ sampling from $\mathcal{N}(0,1)$, and $\sigma$ is the standard deviation of the added noise in each update. This formulation allows the CNN to learn a mapping from the random noise input to the residual between successive frames. 

Further implementation details are provided in the section~\ref{impletment_detail}.

\section{Experiments}

\subsection{Datasets}

Our validation combines video quality assessment and downstream tasks using two datasets: 1) EchoNet-LVH~\cite{echolvh} (1024×768 resolution, 90-200 frames/clip) for restoration evaluation. We simulate LR scenarios through normalization and downsampling with Gaussian noise of various standard deviations (std 0.01-0.1), comparing outputs against original HR videos. 2) EchoNet-Dynamic~\cite{echodynamic} (112×112 resolution, 28-1002 frames/clip) for application testing. We randomly selected 100 videos from each dataset for the respective evaluation scenarios.

\begin{figure}[!t]
\centering
  \includegraphics[height=0.25\linewidth]{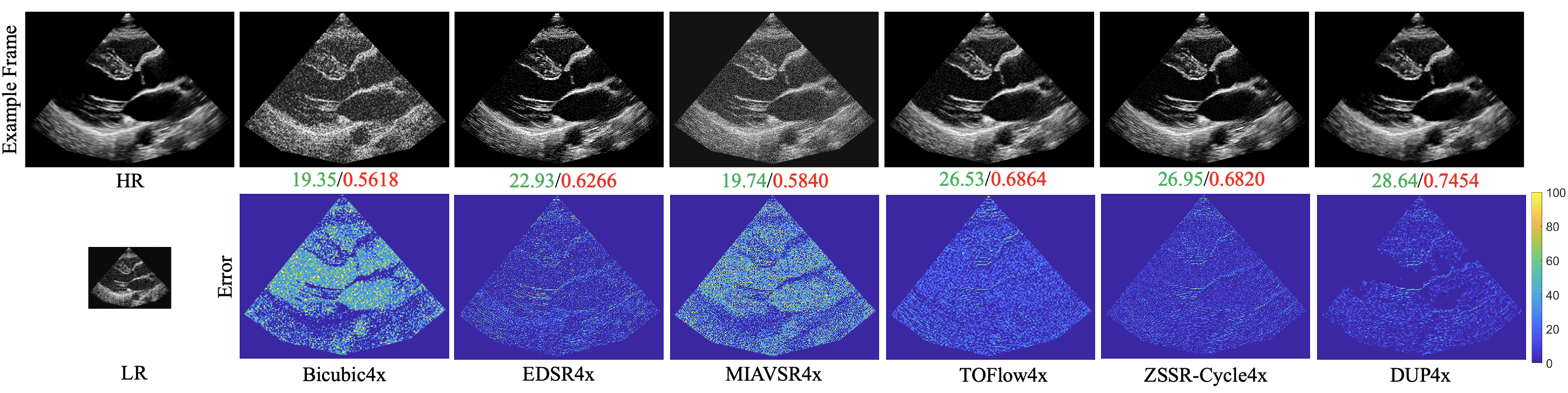}
  \caption{Visual comparisons of $4\times$ SR results. The paired HR and LR example frames are shown on the far left. The upper row displays the SR frames from various methods, with quantitative results (\textcolor{gre}{PSNR (dB)}/\textcolor{red}{SSIM}) shown underneath, while the lower row is the corresponding error maps.}
  \label{vis}
\end{figure}
\subsection{Implement Details}
\label{impletment_detail}
In this part, we explain the implementation of our method and experiments. A Lanczos kernel is used for performing the degradation process $\mathcal{A}$. The CNN is an Encoder-Decoder-style architecture featuring skip concatenation. The architecture consists of four encoder and decoder base units each, which perform convolutions using 128 feature maps, along with batch normalization layers and Leaky ReLU activations. Downsampling is achieved via convolutional layers with a stride of two, while up-sampling is by a Lanczos operator. For the first video frame, we allow up to 3000 iterations, implementing early stopping after 2000 iterations with a patience of 100. For subsequent frames, we reduce the maximum to 2000 iterations, with early stopping commencing at 1000 iterations and patience of 50. $\sigma$ is set to 0.03 and $\lambda_1=\lambda_2=0.01$ in Eq.~(\ref{optim_params}). For the downstream task, we employ EchoCLIP~\cite{christensen2024vision}, a CLIP-like vision-language model trained on medical reports paired with US videos. The restored videos (each frame resized to $224\times 224$) generated by different SR methods are fed into EchoCLIP to perform zero-shot EF prediction. All experiments were conducted in Python 3.8.17 on a PC equipped with Intel\textsuperscript{\textcircled{\textsc{r}}} Xeon\textsuperscript{\textcircled{\textsc{r}}} Silver 4210 Processor CPU 2.20GHz and Nvidia GeForce RTX 3090 GPU with 24G of memory.
\begin{table}[!ht]
\centering
\caption{Ablation studies.}
\begin{tabular}{C{65pt}C{60pt}C{60pt}C{60pt}C{60pt}}
\hline
Method & $HO$ & $TV$ & PSNR (dB)\textuparrow & SSIM\textuparrow \\ \hline
 & \ding{55} & \ding{55} & 25.53 & 0.7133 \\ \multirow{2}{*}{DIP (w/o WIn)} & \checkmark & \ding{55} & 27.16 & 0.7432 \\ & \ding{55} & \checkmark & 28.49 & 0.7432 \\ & \checkmark & \checkmark & 28.63 & 0.7440 \\ \hline
 & \ding{55} & \ding{55} & 25.08 & 0.7069 \\ \multirow{2}{*}{DUP} & \checkmark & \ding{55} & 26.85 & 0.7409 \\ & \ding{55} & \checkmark & 28.85 & 0.7722 \\ & \checkmark & \checkmark & \textbf{28.99} & \textbf{0.7740} \\
\hline
\end{tabular}
\label{ablation}
\end{table}
\begin{figure}[ht]
\centering
  \includegraphics[height=0.2\textheight]{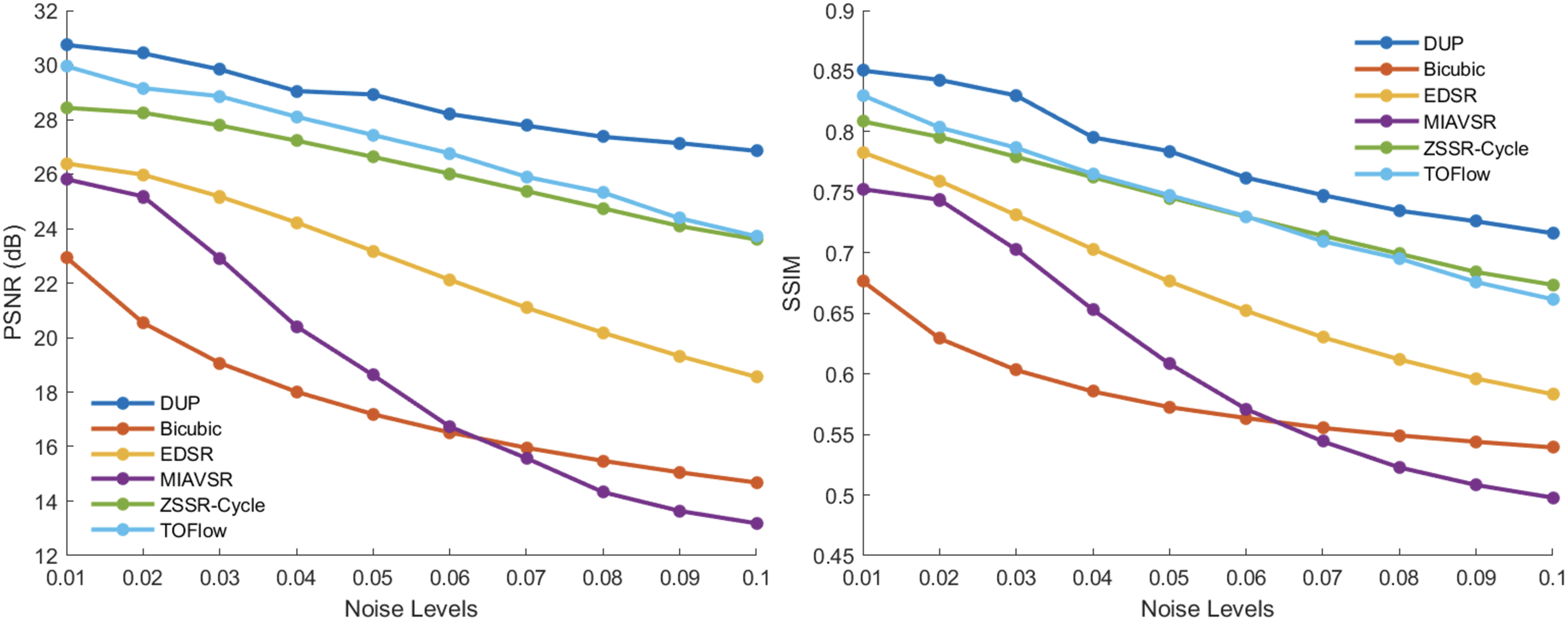}
  \caption{Quantitative evaluations on $4\times$SR videos corrupted by multi-level noise.}
  \label{quant}
\end{figure}
\subsection{Image-based Evaluations}

We compared DUP against four categories of methods: (1) Bicubic interpolation, (2) SISR learning-based architectures (EDSR~\cite{lim2017enhanced} and DIP~\cite{Ulyanov_2018_CVPR}), (3) video SR-based methods (TOFlow~\cite{xue2019video} and MIAVSR~\cite{zhou2024video}), and (4) SR framework for US (ZSSR-Cycle~\cite{liu2021perception}). The visual comparison results in Fig.~\ref{vis}, based on an LR video with $0.05$ std of additive noise, reveal that DUP notably outperforms other methods, specifically in reconstructing clear textures of the ventricles with significantly lower error. In contrast, the results from other methods exhibit noticeable errors after restoration, compromising image clarity. This visual superiority aligns with the quantitative metrics, where DUP attains the highest peak signal-to-noise ratio (PSNR) and structural similarity (SSIM) values. To evaluate the robustness against noise as demonstrated in Fig.~\ref{quant}, DUP consistently achieves the best reconstruction performance under different levels of noise corruption, while other baselines degrade rapidly. From both quantitative and qualitative aspects, DUP yields a superior result with rich details and less noise. In order to analyze the role of regularizations and WIn, we conduct ablation studies on these setups. As listed in Table~\ref{ablation}, the best SNR and structural detail are achieved when both HO and TV priors are incorporated into the target loss, regardless of the presence of WIn. Without WIn, DUP defaults to the same training strategy as DIP, resulting in more iterations(3000 iters/frame) and longer optimization time. Despite this, DUP with the WIn strategy outperforms and achieves greater efficiency.

\subsection{Downstream Evaluations}

\begin{figure}[t]
\centering
  \includegraphics[height=0.2\textheight]{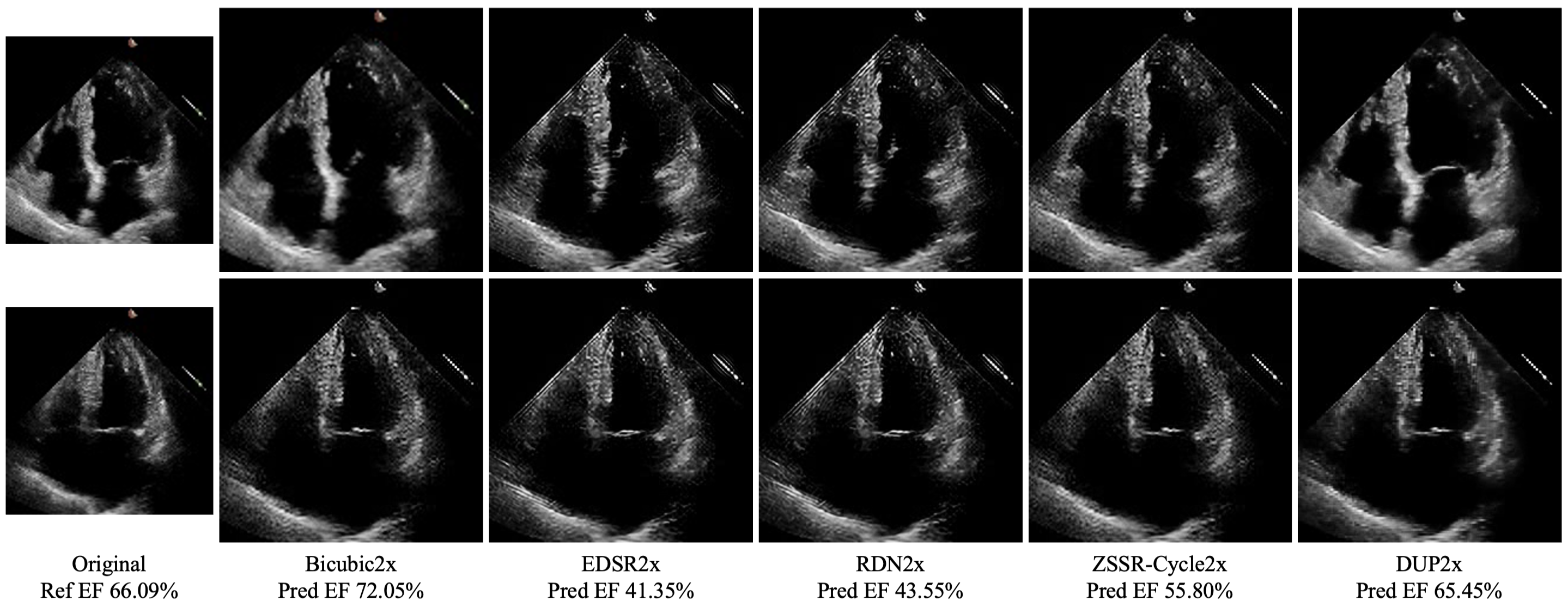}
  \caption{Visualization of maximum diastole (upper) and minimum systole (lower) detected within cardiac cycles. }
  \label{ds_vis}
\end{figure}

To evaluate the practical utility of our proposed method, we assess its performance on the downstream task of EF prediction. Since the EF prediction model only accepts 2× SR results and both MIAVSR and TOFlow are limited to 4×, we substituted these two baselines with RDN~\cite{zhang2018residual} for this part of the study. Table~\ref{tab:ef} compares the mean prediction error of EF across different SR preprocessing methods, where DUP achieves the lowest mean error of $7.23\%$. This demonstrates the potential of DUP in enhancing the quality of US videos for diagnostic tasks. Fig.~\ref{ds_vis} visualizes a sample EF prediction result with detected minimum systole and maximum diastole frames within cardiac cycles. EchoCLIP identified frames corresponding to the two heart phases that align with the manually annotated ground truth from the DUP SR results, while those extracted from other SR videos displayed discrepancies. Consequently, the improvement in visual quality directly correlates with the lower error in EF prediction observed in Fig.~\ref{ds_vis}. The sample further validates the practical applicability of DUP in clinical settings.
\begin{table}[!t]
\centering
\caption{EF Prediction Error}
\label{tab:ef}
\begin{tabular}{c@{\hspace{20pt}} c@{\hspace{20pt}} c@{\hspace{20pt}} c@{\hspace{20pt}} c@{\hspace{20pt}} c@{\hspace{20pt}}}
\hline
SR Method & Bicubic & EDSR & RDN & ZSSR-Cycle & DUP \\ \hline
Mean Error ($\%$) \textdownarrow & 10.26 & 8.69 & 8.46 & 7.60 & \textbf{7.23} \\
\hline
\end{tabular}
\end{table}

\section{Conclusion}
In this work, we propose a self-supervised learning method, Deep Ultrasound Prior (DUP), for US video restoration, eliminating the need for HR video supervision. DUP increases spatial resolution while using CNNs' implicit regularization to capture detailed features accurately. By incorporating the WIn strategy with an early stopping and input update mechanism, the network can perceive and share local frame information while improving convergence speed and performance. With the addition of TV regularization and HO terms, DUP effectively eliminates the measurement error (e.g., noise and artifacts) commonly found in US videos. We demonstrate the superiority of DUP by comparing the quality of super-resolved US videos and their robustness against noise with state-of-the-art methods. Ablation studies confirm the necessity of the WIn strategy and explicit regularization terms for effective video restoration. Furthermore, we validate DUP's capabilities as a pre-processing method in downstream tasks, showing that the restored videos enhance the accuracy of EF predictions for cardiac. This work establishes DUP as a comprehensive solution for US video restoration and overcoming the bottlenecks of its clinical applications.

\begin{credits}
\subsubsection{\ackname} 
This work is partially supported by HKRGC (grant number CityU11301120, C1013-21GF, CityU11309922, CityU9380162), ITF (grant number LU BGR 105824, MHP/054/22), and InnoHK-Hong Kong Centre for Cerebro-cardiovascular Health Engineering.

\subsubsection{\discintname}
The authors have no competing interests to declare that
are relevant to the content of this article.
\end{credits}

%
%
\bibliographystyle{splncs04}  
\bibliography{references}
\end{document}